\documentclass{article}

\usepackage{arxiv}

\usepackage{cite} 
\usepackage{amsmath,amssymb,amsfonts,amsthm}
\usepackage[utf8]{inputenc} 
\usepackage[T1]{fontenc}    
\usepackage[english]{babel}
\usepackage{url}
\usepackage{verbatim}       
\usepackage{adjustbox}      
\usepackage{multirow, tabularx, array, booktabs}        
\usepackage{hyperref}       

\def\BibTeX{{\rm B\kern-.05em{\sc i\kern-.025em b}\kern-.08em
    T\kern-.1667em\lower.7ex\hbox{E}\kern-.125emX}}

\title{Enhancing Generalization in Convolutional Neural Networks through Regularization with Edge and Line Features}

\date{} 					

\author{ 
    \href{https://orcid.org/0000-0002-7039-5189}{\includegraphics[scale=0.06]{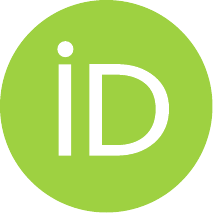}\hspace{1mm}Christoph Linse} \\
	Institute for Neuro- and Bioinformatics\\
	University of L{\"u}beck\\
	L{\"u}beck, Germany \\
	\texttt{c.linse@uni-luebeck.de} \\
	\And
	Beatrice Brückner \\
	Institute for Neuro- and Bioinformatics\\
	University of L{\"u}beck\\
	L{\"u}beck, Germany \\
	\texttt{beatrice.brueckner@t-online.de} \\
    \And
	\href{https://orcid.org/0000-0002-4539-4475}{\includegraphics[scale=0.06]{orcid.pdf}\hspace{1mm}Thomas Martinetz} \\
	Institute for Neuro- and Bioinformatics\\
	University of L{\"u}beck\\
	L{\"u}beck, Germany \\
	\texttt{thomas.martinetz@uni-luebeck.de} \\
}

\renewcommand{\headeright}{}
\renewcommand{\undertitle}{}
\renewcommand{\shorttitle}{Enhancing Generalization with Edge and Line Features}

\hypersetup{
pdftitle={Enhancing Generalization in Convolutional Neural Networks through Regularization with Edge and Line Features},
pdfauthor={Christoph Linse, Beatrice Brückner, Thomas Martinetz},
pdfkeywords={convolutional neural networks, pre-defined filters, edge and line features, regularization},
}

\begin{document}
\maketitle


\begin{abstract}
This paper proposes a novel regularization approach to bias Convolutional Neural Networks (CNNs) toward utilizing edge and line features in their hidden layers. Rather than learning arbitrary kernels, we constrain the convolution layers to edge and line detection kernels. This intentional bias regularizes the models, improving generalization performance, especially on small datasets. As a result, test accuracies improve by margins of $5-11$ percentage points across four challenging fine-grained classification datasets with limited training data and an identical number of trainable parameters.
Instead of traditional convolutional layers, we use Pre-defined Filter Modules, which convolve input data using a fixed set of $3 \times 3$ pre-defined edge and line filters. A subsequent ReLU erases information that did not trigger any positive response. Next, a $1 \times 1$ convolutional layer generates linear combinations. Notably, the pre-defined filters are a fixed component of the architecture, remaining unchanged during the training phase. Our findings reveal that the number of dimensions spanned by the set of pre-defined filters has a low impact on recognition performance. However, the size of the set of filters matters, with nine or more filters providing optimal results.
\end{abstract}


\section{Introduction}
\label{introduction}

Deep Convolutional Neural Networks (CNNs) exhibit strong generalization capabilities on unseen data, especially in image recognition \cite{hertel_deep_2017, linse_large, belkin_2019}. Various methods have emerged to enhance generalization further, including leveraging the power of additional data, transfer learning, or regularization. Our research proposes a new regularization technique to improve the generalization of CNNs by making them utilize edge and line information, two prominent feature types in computer vision. 
In images, edges are boundaries where intensity values change sharply. Therefore, the image gradient conveys information to detect edges. Traditionally, edge detection employs convolution with first-order derivative kernels of various orientations. Similarly, the convolution with second-order derivative kernels can detect thin lines.
While it is known that CNNs can develop such kernels during training, it remains unclear how much they rely on specific features in practice \cite{gavrikov_cnn_2022}. The training data might provide incentives to use other types of information. We demonstrate that the intentional processing of edge and line features in the convolutional layers of CNNs can enhance generalization. We implement this regularization technique by combining understandable pre-defined filters with CNNs. A convolution operation can be described as:
\begin{equation}
    (f * g) [m,n] = \sum_{c=1}^C \sum_{i,j} f_c[i,j] g_c[m-i,n-j].
\end{equation}
Here, $f \in \mathbb{R}^{C \times k \times k}$ is the filter, and $g \in \mathbb{R}^{C \times W \times H}$ is the input feature map with the number of channels $C$, the size of the filter $k$, the width of the image $W$, and its height $H$. 
$m$ and $n$ index the pixels. We express the filters $f_c$ using $k^2$ pre-defined filters $h_1, ... h_{k^2}  \in \mathbb{R}^{1 \times k \times k}$ and weights $w \in \mathbb{R}^{k^2 \times C}$:
\begin{equation}
    f_c[i,j] = \Big(\sum_{l=1}^{k^2} w_{l, c} \cdot h_l\Big) [i,j].
\end{equation}
The convolution becomes:
\begin{equation}
    \text{PFM}_\text{noReLU} [m,n] = (f * g) [m,n] = \sum_{c=1}^C \sum_{l=1}^{k^2} w_{l, c} \cdot (h_l * g_c)[m,n].
    \label{eq:convolution}
\end{equation}
If the set of $h_l$ has a full rank, the network can learn all possible kernels by adjusting the weights $w_{l, c}$. This changes by adding a ReLU \cite{nair_rectified_2010}.
\begin{equation}
    \text{PFM} [m,n] = \sum_{c=1}^C \sum_{l=1}^{k^2} w_{l, c} \cdot \text{ReLU} (h_l * g_c)[m,n]
    \label{eq:pfm_relu}
\end{equation}
The additional ReLU nullifies negative values. It removes the information unrelated to the specific pre-defined filter, leading to a well-structured and comprehensible data representation. The output channels of the ReLU $\text{ReLU} (h_l * g_c)$ form distinct features, each containing positive filter responses only. Later, we choose the $h_l$ as edge and line detectors in different orientations. A subsequent linear combination with the weights $w_{l, c}$ combines the distinct features.
For implementation, we use the Pre-defined Filter Module ($\text{PFM}$), which was initially proposed in the paper \cite{linse_convolutional_2023} to reduce the number of trainable parameters within CNNs. It employs a depthwise $3 \times 3$ convolution, a batch normalization layer (here not shown), a subsequent ReLU, and a pixel-wise $1 \times 1$ convolution.

This research identifies edge and line filters as suitable pre-defined filters for a broad range of vision problems. First, we demonstrate the effectiveness of our regularization method using our own toy dataset for binary image classification. The dataset is available on \url{github.com/Criscraft/Oriented_Dashes_Classification_Dataset}. The task requires counting dashes in different orientations. We show that one $\text{PFM}$ with pre-defined edge and line filters can solve the problem with only two trainable parameters, while a fully convolutional network with $3 \times 3$ kernels and ReLUs would require at least two layers and 36 parameters. Therefore, $\text{PFMs}$ seem to be well-suited for problems where the orientations of edges and lines are relevant.

Second, we show that the generalization abilities of ResNet \cite{he_deep_2016} and DenseNet \cite{huang2017densely} improve on the Fine-Grained Visual Classification of Aircraft dataset (FGVC Aircraft) \cite{maji_fine_grained_2013}, StanfordCars \cite{krause_3d_2013}, Caltech-UCSD Birds-200-2011 (CUB) \cite{wah_caltech_ucsd_2011}, and the 102 Category Flower dataset (Flowers) \cite{nilsback_automated_2008}. 
In additional experiments, we extract random features, achieved through randomly generated pre-defined filters drawn from a uniform distribution around zero. 
This setting does not harm the performance and sometimes increases the test accuracy. The results align with findings in the literature where the power of random filters is discussed \cite{gavrikov_rethinking_2023, ramanujan_whats_2020}.

Third, in a comprehensive overview, we study how many pre-defined filters are needed to reach high recognition rates. Furthermore, we show that the number of dimensions spanned by the set of filters has a minor effect on performance.

The paper is structured as follows.
Section \ref{sec:related_work} describes the related work.
Section \ref{sec:toy_dataset} presents the toy dataset and shows how our regularized architecture solves the dataset with very few parameters.
Implementation details are presented in Section \ref{sec:architecture}.
We show the performance metrics of our approach in Section \ref{sec:benchmarks}.
In Section \ref{sec:dimensions}, we argue why adding linearly dependent edge kernels to the filter set can further increase performance.
Section \ref{sec:number_of_filters} studies how the number of pre-defined filters affects performance.
Section \ref{sec:discussion} discusses the results.


\section{Related Work}
\label{sec:related_work}

Wimmer et al. \cite{wimmerpruning} expressed $3 \times 3$ convolution kernels through various spanning vectors, as detailed in \eqref{eq:convolution}. Their approach, termed interspace pruning, aims to reduce the number of trainable parameters. Our work also employs spanning vectors, but we utilize pre-defined filters that are not adjusted to the training data. Another distinction lies in the intermediate ReLU function in \eqref{eq:pfm_relu} that eliminates patterns to which the filters do not respond positively.

A related study \cite{linse_convolutional_2023} used pre-defined filters to decrease the number of parameters in CNNs. While our approach shares similarities with the related study, our primary objective differs. We aim to regularize CNNs to improve their generalization capabilities rather than only focusing on parameter reduction. We incorporate their $\text{PFMs}$ to improve the performance on four benchmark datasets.
The module parameter $f$ describes the internal number of copies of input channels. The previous study used $f=1$ to save trainable parameters. Here, we choose $f$ as the number of pre-defined filters in the filter pool. For $f=9$ with nine edge- and line filters, the module convolves each input channel using each pre-defined filter as shown in \eqref{eq:pfm_relu}. This setting does not save any parameters when compared to the baseline model.

The $\text{PFM}$ is similar to depthwise separable convolution \cite{howard2017mobilenets}, which also has depthwise and pointwise convolution parts. However, our method does not adjust the filters in the depthwise part during training. Instead, our approach focuses on learning linear combinations of pre-defined filter outputs.


\section{Toy Dataset}
\label{sec:toy_dataset}

\begin{figure}[tb]
	\centering
	\adjustimage{width=0.7\linewidth}{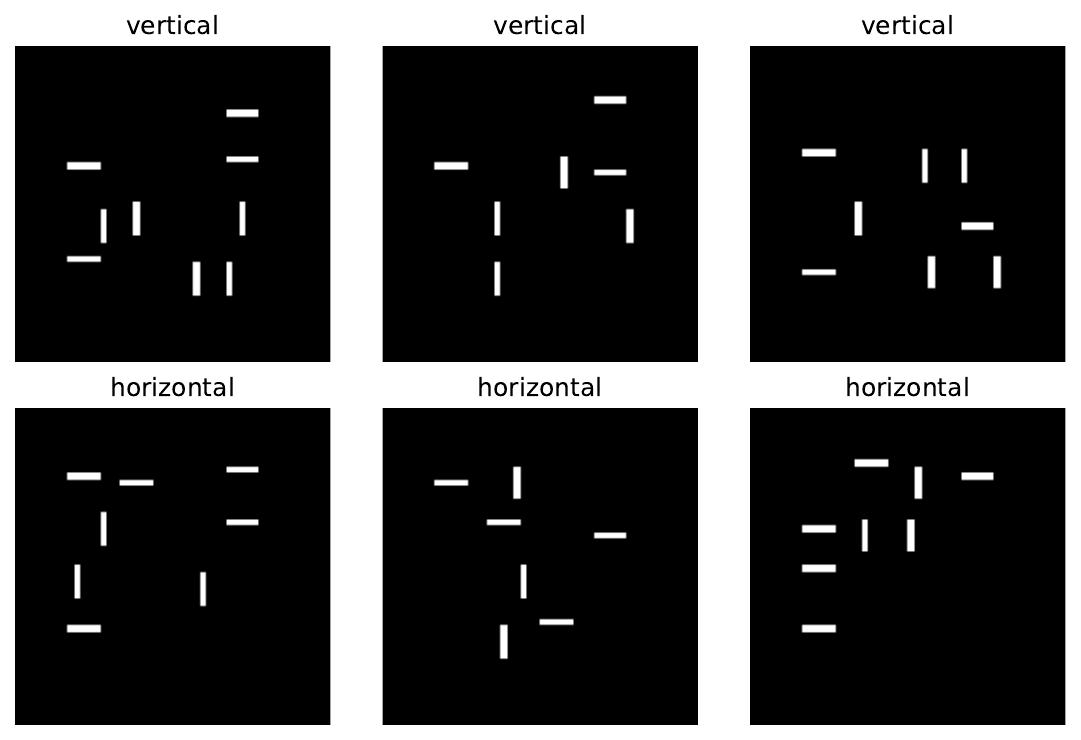}
	\caption{Samples from the toy dataset with the classes \textit{vertical} and \textit{horizontal}.}
    \label{fig:toy_dataset}
\end{figure}

\begin{table}[tb]
\caption{Different implementations of \eqref{eq:toy_dataset_fn}. p denotes the number of trainable parameters. Orange boxes contain trainable parameters. Blue boxes contain pre-defined filters. A dashed line corresponds to the weight value zero. $I^+$ denotes the identity.}
\label{tab:toydataset_architectures}
\centering
{ \scriptsize
\begin{tabular}{ccc}
Model & p & Illustration \\
\midrule
$\text{PFM}$ & 2 & \adjustimage{trim=0.5cm 18cm 5.5cm 0.0cm, clip, width=7cm, valign=m}{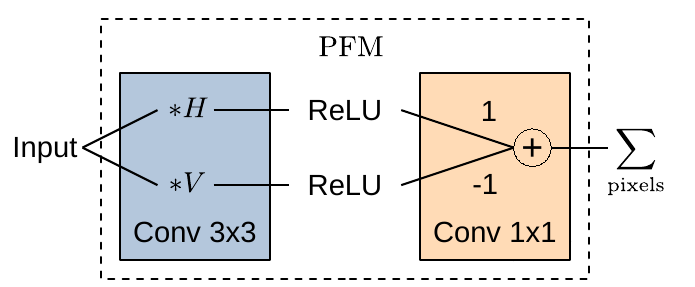} \\
\midrule
CNN & 36 & \adjustimage{trim=0.5cm 13cm 5.5cm 6.0cm, clip, width=7cm, valign=m}{toy_example_architectures.pdf} \\
\midrule
$\text{PFM}_\text{noReLU}$ & 12 & \adjustimage{trim=0.5cm 2.5cm 0.5cm 10.0cm, clip, width=9cm, valign=m}{toy_example_architectures.pdf} \\
\end{tabular}
} 
\end{table}

We study a simplified dataset for binary image classification that requires the processing of gradient information. We demonstrate that pre-defined filters are well-suited to tackle such problems using a minimal number of parameters. The dataset contains 1024 images featuring various horizontal and vertical dashes. The task is determining whether the image has more horizontal than vertical dashes, requiring the network to identify the orientations. Thus, effective utilization of gradient information is essential for solving this problem. Figure \ref{fig:toy_dataset} shows some example images. The grayscale images have $48 \times 48$ pixels. The dashes are one pixel thick and five pixels long. Scenarios with an equal number of horizontal and vertical dashes do not occur. The function $f : \mathbb{R}^{H \times W} \rightarrow \mathbb{R}$
\begin{equation}
f(\mathbf{x}) = \sum_{i, j} \left( \text{ReLU}(H * \mathbf{x}) - \text{ReLU}(V * \mathbf{x}) \right) [i, j]
\label{eq:toy_dataset_fn}
\end{equation}
solves the problem with the pre-defined edge kernels
\begin{equation}
H = 
\begin{pmatrix}
-1 & -1 & -1 \\
2 & 2 & 2 \\
-1 & -1 & -1 
\end{pmatrix} , \quad
V = 
\begin{pmatrix}
-1 & 2 & -1 \\
-1 & 2 & -1 \\
-1 & 2 & -1 
\end{pmatrix} . \\
\end{equation}
A non-negative output refers to the \textit{horizontal} class, while a negative one refers to the \textit{vertical} class. Indeed, \eqref{eq:toy_dataset_fn} correctly classifies all images in the dataset.

Table \ref{tab:toydataset_architectures} presents three variants to implement \eqref{eq:toy_dataset_fn} as a CNN. For the sake of simplicity, we ignore padding in our examples.
The first variant uses our approach. It employs only one $\text{PFM}$ consisting of two pre-defined edge kernels, a ReLU, and a trainable $1 \times 1 $ convolution. Our variant implements \eqref{eq:toy_dataset_fn} with only two trainable parameters.
The second variant uses two common convolutional layers connected by a ReLU. The first convolutional layer has one input and two output channels. The second layer has two input channels and one output channel. The architecture needs a total of 36 trainable parameters.
The third variant employs two $\text{PFM}_\text{noReLU}$ without the intermediate ReLU function. It needs a third pre-defined kernel (the identity) to implement \eqref{eq:toy_dataset_fn}. Here, all layers share the same set of pre-defined filters. The third variant has 12 trainable parameters.

The first variant using $\text{PFM}$ offers the implementation with the fewest trainable parameters. It appears well-suited for image recognition problems where image gradients are relevant. The module enables the network to directly utilize gradient information, effectively filtering out other types of information. In the subsequent section, we apply $\text{PFM}$ in deep CNN architectures and demonstrate that they provide a beneficial bias for challenging real-world image recognition problems.


\section{Architectures and Sets of Filters}
\label{sec:architecture}

Starting from ResNet18 as the baseline, all convolutional layers are substituted with $\text{PFMs}$ \cite{linse_convolutional_2023}. The resulting network is denoted PFNet18. Additionally, ResNet18 contains three skip connections with a $1 \times 1$ convolution and stride two. Following the recommendations of the paper \cite{linse_convolutional_2023}, the skip connections are enhanced by smoothing their inputs using a Gaussian filter. This step addresses aliasing issues and increases the performance of PFNets.

The number of trainable parameters of PFNet18 depends on the number of pre-defined filters as seen in \eqref{eq:pfm_relu}. Table \ref{tab:trainable_parameters} shows the model sizes for different filter sets. PFNet with nine filters exhibits the same number of parameters as the baseline ResNet18. Table \ref{tab:trainable_parameters} also demonstrates the time needed for the forward and backward pass. The times were measured for input tensors of shape $48 \times 3 \times 224 \times 224$ on an NVIDIA GeForce RTX 4090 GPU. Our regularization method tends to be slower compared to the baseline. One reason is that each convolutional layer in the original network is replaced by a PFM with two convolution steps, leading to more nodes in the computational graph.

To further study the applicability of our regularization method, we introduce PFMs to DenseNet121 \cite{huang2017densely} as an additional backbone architecture. Similar to ResNet, DenseNet is a widely used architecture in vision. It consists of 121 layers and contains 12.7 million trainable parameters. DenseNet, or \emph{Densely Connected Convolutional Network}, connects all layers within a dense block in a feedforward manner. Each dense layer in a block receives the feature maps of all preceding layers as input, enhancing feature reuse and improving gradient flow.

\begin{figure}[tb]
	\centering
	\adjustimage{trim=0.5cm 0.5cm 0.5cm 0.5cm, clip, width=0.8\linewidth}{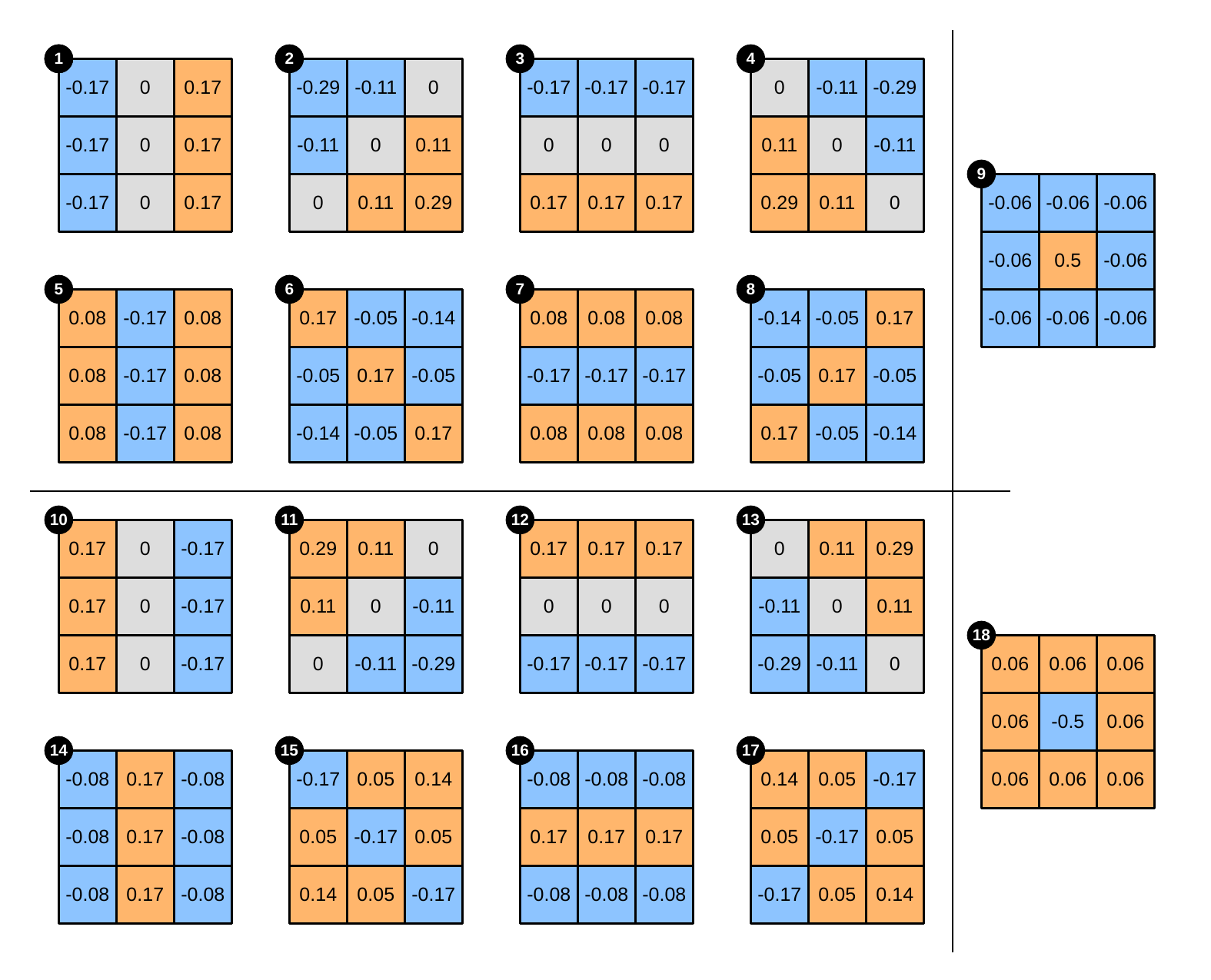}
	\caption{Set of pre-defined filter kernels used in the experiments.}
    \label{fig:filter_kernels}
\end{figure}

All $\text{PFM}$ modules in our experiments share the same pre-defined kernels, which remain unchanged throughout the training process. The experiments involve edge and line detectors in various orientations, as shown in Figure \ref{fig:filter_kernels}. The filters are mean-free, and the sum of their absolute elements is one. We also employ random filters where the filter elements are drawn from a uniform distribution [-1, 1] without normalization. All $\text{PFM}$ modules of the same network share the same random filters. Nevertheless, distinct random seeds lead to different filters. We also utilize translating filters that have one element being one and the other elements being zero.

\begin{table}[tb]
\caption{Number of trainable parameters of PFNet18 in millions. Below, the time for the forward pass (FP) and backward pass (BP) is measured for input tensors of shape $48 \times 3 \times 224 \times 224$ in milliseconds on an NVIDIA GeForce RTX 4090 GPU.}
\label{tab:trainable_parameters}
\centering
\begin{tabular}{cccccccc}
\# Filters:     & 2   & 4     & 8     & 9     & 13    & 18  & ResNet18 \\
\toprule
\# Parameters:  & 2.7 & 5.2   & 10.1  & 11.3  & 16.2  & 22.4 & 11.3\\
Time FP:  & 7 & 11   & 21  & 24  & 33  & 46 & 6 \\
Time BP:  & 14 & 21   & 36  & 40  & 54  & 73 & 14 \\
\end{tabular}
\end{table}


\section{Performance on Benchmark Datasets}
\label{sec:benchmarks}

The models are trained and tested on the ImageNet Large Scale Visual Recognition Challenge (ILSVRC) \cite{russakovsky_imagenet_2015}, Fine-Grained Visual Classification of Aircraft dataset (FGVC Aircraft) \cite{maji_fine_grained_2013}, StanfordCars \cite{krause_3d_2013}, Caltech-UCSD Birds-200-2011 (CUB) \cite{wah_caltech_ucsd_2011}, and the 102 Category Flower dataset (Flowers) \cite{nilsback_automated_2008}.
The CUB dataset has 5994 training and 5794 test images of 200 bird species. The images show birds in natural habitats, posing challenges like variations in lighting and backgrounds.
The FGVC Aircraft dataset includes 6667 training and 3333 test images of 100 airplane models. The classes have notable intra-class variation due to various factors such as advertisement, airlines, and perspective.
The Flowers dataset has 102 blossom types with strong intra-class variations. Regarding the Flowers dataset, we use the union of the official training and validation sets for training consisting of 2040 images. Testing occurs on the official test set with 6149 images.
StanfordCars (2013) has 8144 training and 8041 test images of 196 car models, with images depicting single cars in various environments.
The ILSVRC features 1281167 training images and 50000 validation images across 1000 classes. The images were extracted from various platforms, including Flickr, and were manually labeled with exactly one category.

The trainable network weights are initialized using Kaiming initialization \cite{he_delving_2015}. The training hyperparameters for the ILSVRC are summarized in the Appendix \ref{sec:appendix:training_hyperparameters}. We use the training hyperparameters from the paper \cite{linse_convolutional_2023} for the remaining datasets.
Table \ref{tab:test_performance} presents the average test performance of five models trained with different seeds. The filter type 'edge, line' uses the edge and line detectors from Figure \ref{fig:filter_kernels} starting from index one. The sets with 9 and 18 detectors both span 9 dimensions.

\begin{table}[tb]
\caption{Average test accuracy. The pre-defined filters are not adjusted to the training data.}
\label{tab:test_performance}
\centering
\begin{adjustbox}{width=\linewidth}
\begin{tabular}{ccccccc}
Backbone & Filter type & \# Filters & Flowers    & CUB       & FGVC Aircraft & StanfordCars \\
\toprule
ResNet18 & Translating           & 9       & 73.01$\pm$0.61          & 55.59$\pm$0.62         & 73.72$\pm$0.39           & 77.47$\pm$0.44           \\
ResNet18 & Random & 9       & 78.82$\pm$1.99          & 60.29$\pm$0.62         & 74.59$\pm$3.74           & 79.80$\pm$2.64           \\
ResNet18 & Random & 18      & 81.16$\pm$1.80           & 62.66$\pm$1.01         & 77.60$\pm$1.59           & 81.96$\pm$1.45           \\
ResNet18 & Edge, line   & 9       & 84.28$\pm$0.23          & 61.28$\pm$0.28         & 79.66$\pm$0.20            & 82.64$\pm$0.17           \\
ResNet18 & Edge, line   & 18      & \textbf{85.16$\pm$0.15} & \textbf{63.01$\pm$0.50} & \textbf{81.66$\pm$0.23}  & \textbf{83.66$\pm$0.20}  \\
ResNet18 \cite{linse_convolutional_2023} & -- & --      & 73.4$\pm$0.34          & 58.51$\pm$0.53         & 73.32$\pm$1.06           & 77.9$\pm$0.37           \\
\midrule
DenseNet121 & Edge, line & 9       & 81.46$\pm$0.38          & 60.43$\pm$0.18         & 78.94$\pm$0.29            & 80.62$\pm$0.32           \\
DenseNet121 & Edge, line & 18 & \textbf{81.74$\pm$0.35} & \textbf{62.10$\pm$0.37} & \textbf{80.44$\pm$0.17}  & \textbf{81.04$\pm$0.34}  \\
DenseNet121 & -- & --  & 75.19$\pm$0.66          & 58.26$\pm$0.60         & 74.03$\pm$0.38           & 77.43$\pm$0.26           \\
\end{tabular}
\end{adjustbox}
\end{table}

The edge and line detectors exhibit performance improvements, achieving margins of $5-11$ percentage points compared to the baseline. This enhancement is consistent across all four datasets and is attributed to processing edge and line features in the convolutional layers, contributing to the regularization of the models. The experiments with DenseNet121 as a backbone show similar results. Interestingly, having more than 9 filters enhances the test performance further, even though the additional filters are linearly dependent. Section \ref{sec:dimensions} studies this phenomenon in detail.

Experiments with pre-defined filters, randomly drawn from a uniform distribution around zero, are also shown in Table \ref{tab:test_performance}. Occasionally, PFNet18 with random filters surpasses the baseline model. However, the networks exhibit a high standard deviation in test accuracy, reaching up to $3.74\%$ for the FGVC Aircraft dataset. Apparently, some random filter sets are more or less suited to the specific recognition tasks.

In alternative experiments, we employ translating filters instead of edge, line, or random filters. Referring to \eqref{eq:pfm_relu}, the translating filters mimic the learning of the convolutional filters in the canonical basis. Compared to ResNet18, the performance drops up to $3\%$. We attribute this decline to the ReLU in the first layer, which sets approximately half of the pixels of the original input image to zero. The impact appears to depend on the dataset.

Table \ref{tab:test_performance_imagenet} presents the test accuracies on the ILSVRC (ImageNet). Our regularized model with nine filters exhibits a performance that is $2\%$ below the baseline. We attribute the observed decline in performance to the similarity between training and test accuracy. The similarity implies that regularization might not be necessary in this scenario. Instead, the model's capacity should ideally increase. Indeed, when we augment the network with more filters, thereby expanding its capacity, the model shows a slight improvement over the baseline.

\begin{table}[tb]
\caption{Accuracy on the validation set of the ILSVRC (ImageNet). The pre-defined filters are not adjusted to the training data.}
\label{tab:test_performance_imagenet}
\centering
\begin{tabular}{cccc}
Filter type          & \# Filters & Top 1 & Top 5  \\
\toprule
ResNet18 \cite{linse_convolutional_2023}    & --     & 69.60     & 89.13 \\
Edge, line      & 9     & 67.59     & 88.13 \\
Edge, line      & 18    & \textbf{70.16}     & \textbf{89.49} \\
\end{tabular}
\end{table}

Furthermore, we adjust the pre-defined filters to the train data. We allow each $\text{PFM}$ module to learn its own set with nine or 18 filters. However, these experiments do not improve the performance metrics, as shown in Table \ref{tab:test_performance_trainable_filters}. The training process struggled to identify filters that outperformed our pre-defined edge and line detectors, underscoring their good generalization abilities.

\begin{table}[tb]
\caption{Average test accuracy. The first column describes the initialization of the pre-defined filters. The filters are adjusted to the training data.}
\label{tab:test_performance_trainable_filters}
\centering
\begin{adjustbox}{width=0.8\linewidth}
\begin{tabular}{cccccc}
Filter type                    & \# Filters & Flowers    & CUB       & FGVC Aircraft & StanfordCars \\
\toprule
Translating  & 9        & 74.22$\pm$0.92          & 59.35$\pm$0.80         & 74.56$\pm$0.38           & 79.29$\pm$0.70           \\
Random & 9              & 79.61$\pm$1.96          & 60.92$\pm$1.58         & 75.19$\pm$3.25           & 80.08$\pm$2.31           \\
Random & 18             & 80.65$\pm$1.35           & 62.94$\pm$0.89         & 78.01$\pm$1.89           & 81.81$\pm$1.61           \\
Edge, line   & 9          & 84.29$\pm$0.33          & 61.93$\pm$0.49         & 78.09$\pm$0.36            & 81.54$\pm$0.45           \\
Edge, line   & 18         & \textbf{84.62$\pm$0.41} & \textbf{63.17$\pm$0.48} & \textbf{79.54$\pm$0.55}  & \textbf{82.74$\pm$0.27} \\
ResNet18 \cite{linse_convolutional_2023}                & --      & 73.4$\pm$0.34          & 58.51$\pm$0.53         & 73.32$\pm$1.06           & 77.9$\pm$0.37           \\
\end{tabular}
\end{adjustbox}
\end{table}

\section{Limited Impact of the Spanned Dimensions on Performance}
\label{sec:dimensions}

\begin{figure}[tbp]
	\centering
	\adjustimage{width=0.7\linewidth}{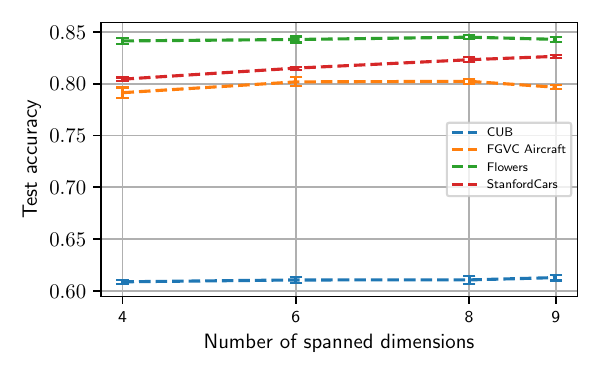}
	\caption{Average test accuracy when using nine edge and line detectors that span a variable number of dimensions.}
    \label{fig:plot_double_relu_pfm_variable_rank}
\end{figure}

This section investigates how the number of dimensions spanned by the set of pre-defined filters affects the recognition performance. The number of dimensions means how many of the nine possible dimensions are spanned by the set of pre-defined filters. It is obtained by counting the number of linearly independent filters. PFNet18 is trained using nine pre-defined kernels that span a varying number of dimensions. To get four dimensions, we choose the filters 1, 3, 5, 7, 10, 12, 14, and 16, and the sum of 14 and 16 from Figure \ref{fig:filter_kernels}. Figure \ref{fig:plot_double_relu_pfm_variable_rank} presents the average test accuracies from five runs with different seeds. The results suggest that the dimensionality spanned by the pre-defined filters has minimal influence on performance, with four dimensions already yielding satisfactory results.

Interestingly, Table \ref{tab:test_performance} unveils a notable performance gain between using 9 and 18 pre-defined kernels by margins of $1-3\%$. The nine $3 \times 3$ pre-defined filters already span all 9 dimensions. We suggest that the additional ReLU in the $\text{PFM}$ can make the features linearly independent, even if the filter kernels are linearly dependent. A module with one input channel and two pre-defined filters $\tilde{w}_1, \tilde{w}_2 \in \mathbb{R}^{k \times k}$ contains the two functions
\begin{equation}
\begin{split}
    f^{(\tilde{w}_1, m, n)}(\mathbf{x}) &= \text{ReLU}(\tilde{w}_1 * \mathbf{x})[m,n] \\
    f^{(\tilde{w}_2, m, n)}(\mathbf{x}) &= \text{ReLU}(\tilde{w}_2 * \mathbf{x})[m,n] \\
    f^{(\tilde{w}_1, m, n)}&, f^{(\tilde{w}_2, m, n)} : \mathbb{R}^{M \times N} \rightarrow \mathbb{R} \\
\end{split}
\label{eq:lin_independence_relu}
\end{equation}
with pixel coordinates $m, n \in N$. As shown in the appendix \ref{sec:appendix:linear_independency}, if $a \tilde{w}_1 = \tilde{w}_2$ with $a < 0$, then the functions are linearly independent.
The $\text{PFM}$ module benefits from having a negative copy of a pre-defined filter because the rectified convolution outputs become linearly independent. The entire $\text{PFM}$ module (ignoring normalization layers) can be written as
\begin{equation}
\begin{split}
    \text{PFM} [m,n] &= \sum_{c=1}^{c_{\text{in}}} \sum_{l=1}^{2} q_{cl} \text{ReLU}(\tilde{w}_l * \mathbf{x})[m,n] \\
    &= q_{11} \text{ReLU}(\tilde{w}_1 * \mathbf{x})[m,n] \\
    & \quad + q_{12} \text{ReLU}(a \tilde{w}_1 * \mathbf{x})[m,n] \\
    \text{Case 1: } &(\tilde{w}_1 * \mathbf{x})[m,n] \geq 0 : q_{11} (\tilde{w}_1 * \mathbf{x})[m,n]\\
    \text{Case 2: } &(\tilde{w}_1 * \mathbf{x})[m,n] < 0 : a q_{12} (\tilde{w}_1 * \mathbf{x})[m,n] .\\
\end{split}
\label{eq:lin_independence_relu2}
\end{equation}
The convolution output is either multiplied with $q_{11}$ or $a q_{12}$. Here, ReLU acts like a switch, deciding which weight to apply.

We conclude that the set of pre-defined filters should incorporate pairs of filter kernels with inverted signs. For instance, the filters $\tilde{w}_1$ and $\tilde{w}_2 = -\tilde{w}_1$ could represent two edge detectors of opposing directions (see filters one and ten in Figure \ref{fig:filter_kernels}). If the input contains an edge aligned with $\tilde{w}_1$, the first output channel has a positive activation while the second channel remains inactive. If the input contains the opposing edge, the second channel activates while the first channel remains inactive.

To better understand the benefit of linearly dependent filters, we repeat the experiments conducted in Section \ref{sec:benchmarks} in an ablation study. This time, we employ $\text{PFM}_\text{noReLU}$ modules without the additional ReLU function as described in \eqref{eq:convolution}. Since the pre-defined filters span all nine dimensions, the network retains its ability to learn all convolution kernels. Regularization does not occur. As expected, the results presented in Table \ref{tab:test_performance_without_relu} exhibit a notable performance drop for edge and line filters and a weak drop for random kernels. The baseline model ResNet18 outperforms the edge and line filters on the CUB, FGVC Aircraft, and StanfordCars datasets.

Furthermore, Table \ref{tab:test_performance_without_relu} shows that translating filters achieve test accuracies comparable to the baseline. This complements the prior experiments where an additional ReLU function decreased the recognition rates by $3\%$. The additional ReLU decreases the performance by setting dark input pixels to zero, erasing half of the original image's information.

\begin{table}[tb]
\caption{Average test performance on the benchmark datasets. All experiments but the baseline use $\text{PFM}_\text{noReLU}$ modules without the additional ReLU. The pre-defined filters are not adjusted to the training data.}
\label{tab:test_performance_without_relu}
\centering
\begin{adjustbox}{width=0.8\linewidth}
\begin{tabular}{cccccc}
Filter type                    & \# Filters & Flowers    & CUB       & FGVC Aircraft & Stanford Cars \\
\toprule
Translating           & 9       & 74.93$\pm$0.72          & 59.62$\pm$0.52         & 74.51$\pm$0.58           & 79.84$\pm$0.40           \\
Random & 9       & 78.51$\pm$1.64          & 59.28$\pm$1.55         & 73.62$\pm$4.21           & 79.11$\pm$2.53           \\
Random & 18      & 78.54$\pm$2.27           & \textbf{60.55$\pm$1.02}         & \textbf{75.05$\pm$2.13}           & \textbf{80.04$\pm$1.47}           \\
Edge, line   & 9       & 78.29$\pm$0.38          & 50.01$\pm$0.69         & 71.25$\pm$0.40            & 72.56$\pm$0.28           \\
Edge, line   & 18      & \textbf{79.38$\pm$0.35} & 50.41$\pm$0.46 & 72.43$\pm$0.59  & 73.68$\pm$0.49 \\
ResNet18 \cite{linse_convolutional_2023}                & --      & 73.4$\pm$0.34          & 58.51$\pm$0.53         & 73.32$\pm$1.06           & 77.9$\pm$0.37           \\
\end{tabular}
\end{adjustbox}
\end{table}

\section{Nine or More Filters are Needed for Optimal Results}
\label{sec:number_of_filters}

\begin{figure}[tb]
	\centering
	\adjustimage{width=0.48\linewidth}{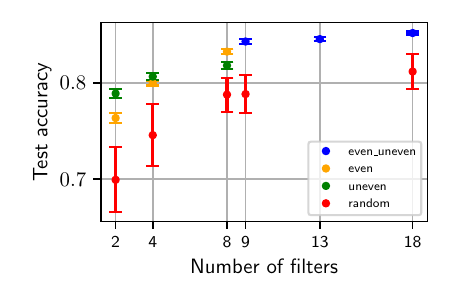}
	\adjustimage{width=0.48\linewidth}{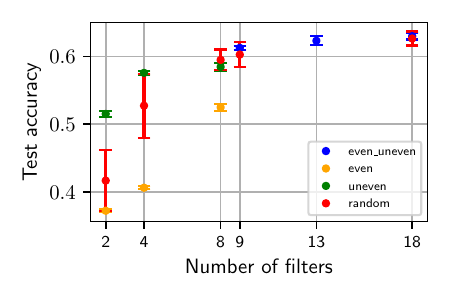}
	\caption{Average test accuracy when using a variable number of filters. Left: Flowers dataset. Right: CUB dataset.}
    \label{fig:plot_double_relu_pfm_variable_number_of_filters}
\end{figure}

This section studies the effect of the number of pre-defined filters on the recognition performance. We train PFNet18 on the CUB and the Flowers dataset using the filter subsets in Table \ref{tab:filter_subsets}. As illustrated in Figure \ref{fig:plot_double_relu_pfm_variable_number_of_filters}, the best results are obtained when utilizing all 18 filters. The test accuracies drop when choosing four or fewer filters. It is worth noting that a reduction in the number of pre-defined filters not only limits the information transferred to the subsequent layer but also leads to a decrease in trainable parameters, thereby diminishing the model's capacity, as shown in Table \ref{tab:trainable_parameters}.

\begin{table}[tb]
\caption{Filter subsets of different sizes and types.}
\label{tab:filter_subsets}
\centering
\begin{adjustbox}{width=0.5\linewidth}
\begin{tabular}{ccc}
Filter type     & \# Filters     & Filter selection (see Fig. \ref{fig:filter_kernels})  \\
\toprule
even            & 2             & 5, 7 \\
even            & 4             & 5, 7, 14, 16 \\
even            & 8             & 5 - 8, 14 - 17 \\
uneven          & 2             & 1, 3 \\
uneven          & 4             & 1, 3, 10, 12 \\
uneven          & 8             & 1 - 4, 10 - 13 \\
even-uneven     & 9             & 1 - 9 \\
even-uneven     & 13            & 1 - 9, 11, 13, 15, 17 \\
even-uneven     & 18            & 1 - 18 \\
\end{tabular}
\end{adjustbox}
\end{table}

Interestingly, the edge filters (green color) often outperform the line filters (yellow color) or the random filters (red color). Given the abundance of edges in images, we hypothesize that edge filters convey more information than lines.


\section{Discussion}
\label{sec:discussion}

Across four fine-grained classification datasets, we observed a notable test accuracy improvement ranging between $5-11$ percentage points using nine edge and line filters while maintaining the same number of parameters as the baseline model, ResNet18. Doubling the number of pre-defined filters resulted in further performance improvements. We also applied our regularization method to DenseNet121 with similar results. The experiments demonstrate the beneficial bias introduced by our regularization method. Notably, regularization was not achieved by reducing the number of trainable parameters but by biasing the CNN to process understandable edge and line features.

The ReLU in $\text{PFMs}$ can remove specific information from the incoming feature maps. A question for future research is which pre-defined filters allow or do not allow the learning of the identity mapping. Consequently, we applied our regularization method to models with residual connections (ResNet) and densely connected layers (DenseNet), where the identity mapping is a fundamental part of the architecture.

The number of dimensions spanned by the set of pre-defined filters appears to have a low impact on recognition performance. This observation is attributed to the nature of the ReLU activation function. When applied to the outputs of convolution operations with linearly dependent filter kernels, ReLU can produce linearly independent results. The experiments indicate that four dimensions achieve comparable recognition values to those obtained with nine dimensions.

However, the number of pre-defined filters matters with optimal results obtained with nine or more $3 \times 3$ filters. Unfortunately, using more filters is bound to higher computational costs, as seen in Table \ref{tab:trainable_parameters}, limiting the attractiveness for platforms with sensitive energy and speed requirements.


\section{Conclusion}
\label{sec:conclusion}
Processing edge and line features within CNNs improves their generalization abilities in image recognition. For ResNet18 and DenseNet121, we observed a noteworthy increase in test accuracy from $5-11$ percentage points across four classification benchmark datasets with the same number of trainable parameters. The results imply that pre-defined edge and line filters add a suitable bias to many image recognition problems.
We demonstrated that the number of dimensions spanned by the set of pre-defined filters has a minimal impact on performance. However, the number of pre-defined filters matters. Using fewer than nine pre-defined $3 \times 3$ filters reduces test accuracy while using more than nine filters improves recognition performance but increases computational costs. 

We believe pre-defined filters in CNNs are an underestimated area, offering improved generalization and the potential to save parameters. However, determining the optimal set of pre-defined filters for specific image recognition tasks remains challenging. Better problem-specific filters might exist. Adjusting the filters to the training data did not yield further improvements. Applying pre-defined filters to diverse tasks beyond image recognition, such as medical image, sound, or video analysis, is left for future research. Specialized features may offer significant benefits in these domains.
Furthermore, investigating the compatibility of our regularization method with architectures beyond ResNet and DenseNet requires more experiments. Assessing $\text{PFM}$ in different CNN architectures will help to determine their generalizability and effectiveness.
Future research should also explore transfer learning with pre-defined filters, potentially reducing the need for extensive retraining.


\appendix
\section{Appendix}
\label{sec:appendix}

\subsection{Training Hyperparameters for ImageNet}
\label{sec:appendix:training_hyperparameters}

The networks are trained with a batch size of 48 on five NVIDIA GeForce RTX 4090 GPUs. The remaining training hyperparameters are taken from the training reference of PyTorch \cite{paszke_pytorch_2019}. The cross-entropy loss is minimized using stochastic gradient descent for 90 epochs with a momentum of 0.9 and weight-decay 0.0001. The initial learning rate of 0.1 is reduced by a factor of 0.1 every 30 epochs.

\subsection{Linear Independency of ReLU-based Functions}
\label{sec:appendix:linear_independency}

Two functions $f_1, f_2 : X \rightarrow Y$ are linearly independent if
\begin{equation}
    (\forall \mathbf{x} \in X: c_1 f_1(\mathbf{x}) + c_2 f_2(\mathbf{x}) = 0) \Leftrightarrow c_1 = c_2 = 0 .
\end{equation}
Consider the functions from \eqref{eq:lin_independence_relu} that occur in the $\text{PFM}$ module. The functions are linearly dependent if the pre-defined filters $\tilde{w}_1$ and $\tilde{w}_2$ are linearly dependent and $a \tilde{w}_1 = \tilde{w}_2, a \geq 0$.

\begin{proof}
Choose some arbitrary $\mathbf{x} \in \mathbb{R}^{M \times N}$.
Choose $c_1 \in \mathbb{R} \backslash \{0\}$ and $c_2 = - c_1 / a$. Then, 
\begin{equation}
\begin{split}
    &c_1 f^{(\tilde{w}_1, m, n)}(\mathbf{x}) + c_2 f^{(\tilde{w}_2, m, n)}(\mathbf{x}) \\
    &= c_1 \text{ReLU}(\tilde{w}_1 * \mathbf{x})[m,n] - c_1 \frac{a}{a} \text{ReLU}(\tilde{w}_1 * \mathbf{x})[m,n] = 0 . \\
\end{split}
\end{equation}
\end{proof}

The functions in \eqref{eq:lin_independence_relu} are linearly independent if $\tilde{w}_1$ and $\tilde{w}_2$ are linearly dependent and $a \tilde{w}_1 = \tilde{w}_2, a < 0$.

\begin{proof}
The $\leftarrow$ direction is clear. To show the $\rightarrow$ direction, let $\mathbf{x} \in \mathbb{R}^{M \times N}$. 
\begin{equation}
\begin{split}
    &c_1 f^{(\tilde{w}_1, m, n)}(\mathbf{x}) + c_2 f^{(\tilde{w}_2, m, n)}(\mathbf{x}) = 0 \\
    &\Leftrightarrow c_1 \text{ReLU}(\tilde{w}_1 * \mathbf{x})[m,n] + c_2 \text{ReLU}(a \tilde{w}_1 * \mathbf{x})[m,n] = 0 \\
    &\Leftrightarrow c_1 \text{ReLU}(\tilde{w}_1 * \mathbf{x})[m,n] - a c_2 \text{ReLU}(-\tilde{w}_1 * \mathbf{x})[m,n] = 0 \\
    &\text{Case1}: (\tilde{w}_1 * \mathbf{x})[m,n] \geq 0 \implies c_1 = 0\\
    &\text{Case2}: (\tilde{w}_1 * \mathbf{x})[m,n] < 0 \implies c_2 = 0\\
\end{split}
\end{equation}
The sum has to be zero for all $\mathbf{x} \in \mathbb{R}^{M \times N}$. This means that both coefficients $c_1$ and $c_2$ have to be zero.
\end{proof}


\section*{Acknowledgment}
The work of Christoph Linse was supported by the Bundesministerium f{\"u}r Wirtschaft und Klimaschutz through the Mittelstand-Digital Zentrum Schleswig-Holstein Project.
The Version of Record of this contribution is published in Artificial Neural Networks and Machine Learning – ICANN 2024, Lecture Notes in Computer Science, vol 15016. Springer,
and is available online at \url{https://doi.org/10.1007/978-3-031-72332-2_28}.

\bibliographystyle{ieeetr}
\bibliography{References}

\begin{thebibliography}{10}

\bibitem{hertel_deep_2017}
L.~Hertel, E.~Barth, T.~Käster, and T.~Martinetz, ``Deep {Convolutional} {Neural} {Networks} as {Generic} {Feature} {Extractors},'' {\em arXiv:1710.02286 [cs]}, Oct. 2017.
\newblock arXiv: 1710.02286.

\bibitem{linse_large}
C.~Linse and T.~Martinetz, ``Large neural networks learning from scratch with very few data and without explicit regularization,'' in {\em Proceedings of the 2023 15th International Conference on Machine Learning and Computing}, ICMLC '23, (New York, NY, USA), p.~279–283, Association for Computing Machinery, 2023.

\bibitem{belkin_2019}
M.~Belkin, D.~Hsu, S.~Ma, and S.~Mandal, ``Reconciling modern machine-learning practice and the classical bias--variance trade-off,'' {\em Proceedings of the National Academy of Sciences}, vol.~116, no.~32, pp.~15849--15854, 2019.

\bibitem{gavrikov_cnn_2022}
P.~Gavrikov and J.~Keuper, ``{CNN} {Filter} {DB}: {An} {Empirical} {Investigation} of {Trained} {Convolutional} {Filters},'' in {\em 2022 {IEEE}/{CVF} {Conference} on {Computer} {Vision} and {Pattern} {Recognition} ({CVPR})}, (New Orleans, LA, USA), pp.~19044--19054, IEEE, June 2022.

\bibitem{nair_rectified_2010}
V.~Nair and G.~E. Hinton, ``Rectified {Linear} {Units} {Improve} {Restricted} {Boltzmann} {Machines},'' in {\em Proceedings of the 27th {International} {Conference} on {International} {Conference} on {Machine} {Learning}}, {ICML}'10, (Madison, WI, USA), pp.~807--814, Omnipress, 2010.
\newblock event-place: Haifa, Israel.

\bibitem{linse_convolutional_2023}
C.~Linse, E.~Barth, and T.~Martinetz, ``Convolutional {Neural} {Networks} {Do} {Work} with {Pre}-{Defined} {Filters},'' in {\em 2023 {International} {Joint} {Conference} on {Neural} {Networks} ({IJCNN})}, pp.~1--8, IEEE, 2023.

\bibitem{he_deep_2016}
K.~He, X.~Zhang, S.~Ren, and J.~Sun, ``Deep {Residual} {Learning} for {Image} {Recognition},'' in {\em 2016 {IEEE} {Conference} on {Computer} {Vision} and {Pattern} {Recognition} ({CVPR})}, (Las Vegas, NV, USA), pp.~770--778, IEEE, June 2016.

\bibitem{huang2017densely}
G.~Huang, Z.~Liu, L.~Van Der~Maaten, and K.~Q. Weinberger, ``Densely connected convolutional networks,'' in {\em Proceedings of the IEEE Conference on Computer Vision and Pattern Recognition}, pp.~4700--4708, 2017.

\bibitem{maji_fine_grained_2013}
S.~Maji, E.~Rahtu, J.~Kannala, M.~Blaschko, and A.~Vedaldi, ``Fine-{Grained} {Visual} {Classification} of {Aircraft},'' {\em arXiv:1306.5151 [cs]}, June 2013.
\newblock arXiv: 1306.5151.

\bibitem{krause_3d_2013}
J.~Krause, M.~Stark, J.~Deng, and L.~Fei-Fei, ``{3D} {Object} {Representations} for {Fine}-{Grained} {Categorization},'' in {\em 2013 {IEEE} {International} {Conference} on {Computer} {Vision} {Workshops}}, (Sydney, Australia), pp.~554--561, IEEE, Dec. 2013.

\bibitem{wah_caltech_ucsd_2011}
C.~Wah, S.~Branson, P.~Welinder, P.~Perona, and S.~Belongie, ``The {Caltech}-{UCSD} {Birds}-200-2011 {Dataset},'' Tech. Rep. CNS-TR-2011-001, California Institute of Technology, 2011.

\bibitem{nilsback_automated_2008}
M.-E. Nilsback and A.~Zisserman, ``Automated {Flower} {Classification} over a {Large} {Number} of {Classes},'' in {\em 2008 {Sixth} {Indian} {Conference} on {Computer} {Vision}, {Graphics} \& {Image} {Processing}}, (Bhubaneswar, India), pp.~722--729, IEEE, Dec. 2008.

\bibitem{gavrikov_rethinking_2023}
P.~Gavrikov and J.~Keuper, ``Rethinking 1x1 {Convolutions}: {Can} we train {CNNs} with {Frozen} {Random} {Filters}?,'' Jan. 2023.
\newblock arXiv:2301.11360 [cs].

\bibitem{ramanujan_whats_2020}
V.~Ramanujan, M.~Wortsman, A.~Kembhavi, A.~Farhadi, and M.~Rastegari, ``What's {Hidden} in a {Randomly} {Weighted} {Neural} {Network}?,'' Mar. 2020.
\newblock arXiv:1911.13299 [cs].

\bibitem{wimmerpruning}
P.~Wimmer, J.~Mehnert, and A.~Condurache, ``Interspace pruning: Using adaptive filter representations to improve training of sparse cnns,'' in {\em Proceedings of the IEEE/CVF conference on computer vision and pattern recognition}, pp.~12527--12537, 2022.

\bibitem{howard2017mobilenets}
A.~G. Howard, M.~Zhu, B.~Chen, D.~Kalenichenko, W.~Wang, T.~Weyand, M.~Andreetto, and H.~Adam, ``Mobilenets: Efficient convolutional neural networks for mobile vision applications,'' {\em arXiv preprint arXiv:1704.04861}, 2017.

\bibitem{russakovsky_imagenet_2015}
O.~Russakovsky, J.~Deng, H.~Su, J.~Krause, S.~Satheesh, S.~Ma, Z.~Huang, A.~Karpathy, A.~Khosla, M.~Bernstein, A.~C. Berg, and L.~Fei-Fei, ``{ImageNet} {Large} {Scale} {Visual} {Recognition} {Challenge},'' {\em International Journal of Computer Vision}, vol.~115, pp.~211--252, Dec. 2015.

\bibitem{he_delving_2015}
K.~He, X.~Zhang, S.~Ren, and J.~Sun, ``Delving {Deep} into {Rectifiers}: {Surpassing} {Human}-{Level} {Performance} on {ImageNet} {Classification},'' in {\em 2015 {IEEE} {International} {Conference} on {Computer} {Vision} ({ICCV})}, (Santiago, Chile), pp.~1026--1034, IEEE, Dec. 2015.

\bibitem{paszke_pytorch_2019}
A.~Paszke, S.~Gross, F.~Massa, A.~Lerer, J.~Bradbury, G.~Chanan, T.~Killeen, Z.~Lin, N.~Gimelshein, L.~Antiga, A.~Desmaison, A.~Kopf, E.~Yang, Z.~DeVito, M.~Raison, A.~Tejani, S.~Chilamkurthy, B.~Steiner, L.~Fang, J.~Bai, and S.~Chintala, ``{PyTorch}: {An} {Imperative} {Style}, {High}-{Performance} {Deep} {Learning} {Library},'' in {\em Advances in {Neural} {Information} {Processing} {Systems} 32} (H.~Wallach, H.~Larochelle, A.~Beygelzimer, F.~d. Alché-Buc, E.~Fox, and R.~Garnett, eds.), pp.~8024--8035, Curran Associates, Inc., 2019.

\end{thebibliography}

\end{document}